\pdfoutput=1

\documentclass[11pt]{article}

\usepackage{acl}
\usepackage{booktabs}
\usepackage{multirow}
\usepackage{enumitem}
\usepackage{graphics}
\usepackage{times}
\usepackage{latexsym}
\usepackage{graphicx}
\usepackage[T1]{fontenc}

\usepackage[utf8]{inputenc}

\usepackage{microtype}
\usepackage{amsfonts}
\usepackage{amsmath}
\usepackage{bm}
\usepackage{url}
\usepackage{enumitem}
\usepackage{listings}
\usepackage{makecell}
\title{TextPruner: A Model Pruning Toolkit for Pre-Trained Language Models}

\author{
Ziqing Yang$^\dag$,
Yiming Cui$^\ddag$$^\dag$,
Zhigang Chen$^\dag$ \\
{$^\dag$State Key Laboratory of Cognitive Intelligence, iFLYTEK Research, China} \\
{$^\ddag$Research Center for Social Computing and Information Retrieval (SCIR), } \\
{Harbin Institute of Technology, Harbin, China} \\
$^\dag$\tt\{zqyang5,ymcui,zgchen\}@iflytek.com \\
$^\ddag$\tt ymcui@ir.hit.edu.cn}

\begin{document}
\maketitle
\begin{abstract}
Pre-trained language models have been prevailed in natural language processing and become the backbones of many NLP tasks, but the demands for computational resources have limited their applications. In this paper, we introduce TextPruner, an open-source model pruning toolkit designed for pre-trained language models, targeting fast and easy model compression. TextPruner offers structured post-training pruning methods, including vocabulary pruning and transformer pruning, and can be applied to various models and tasks. We also propose a self-supervised pruning method that can be applied without the labeled data. Our experiments with several NLP tasks demonstrate the ability of TextPruner to reduce the model size without re-training the model. \footnote{The source code and the documentation are available at \url{http://textpruner.hfl-rc.com}}

\end{abstract}

\section{Introduction}

Large pre-trained language models (PLMs) \cite{devlin-etal-2019-bert,liu2019roberta} have achieved great success in a variety of NLP tasks. However, it is difficult to deploy them for real-world applications where computation and memory resources are limited. Reducing the pre-trained model size and speeding up the inference have become a critical issue.

Pruning is a common technique for model compression. It identifies and removes redundant or less important neurons from the networks. From the view of the model structure, pruning methods can be categorized into \textit{unstructured pruning} and \textit{structured pruning}. In the unstructured pruning, each model parameter is individually removed if it reaches some criteria based on the magnitude or importance score \cite{DBLP:journals/corr/HanPTD15, DBLP:conf/iclr/ZhuG18, DBLP:conf/nips/Sanh0R20}. The unstructured pruning results in sparse matrices and allows for significant model compression, but the inference speed can hardly be improved without specialized devices. While in the structured pruning, rows or columns of the parameters are removed from the weight matrices \cite{DBLP:journals/corr/abs-1910-06360,DBLP:conf/nips/MichelLN19,voita-etal-2019-analyzing,lagunas-etal-2021-block,DBLP:conf/nips/HouHSJCL20}. Thus, the resulting model speeds up on the common CPU and GPU devices.

Pruning methods can also be classified into optimization-free methods \cite{DBLP:conf/nips/MichelLN19} and the ones that involve optimization \cite{frankle2018the, lagunas-etal-2021-block}. The latter usually achieves higher performance, but the former runs faster and is more convenient to use.

Pruning PLMs has been of growing interest. Most of the works focus on reducing transformer size while ignoring the vocabulary \cite{abdaoui-etal-2020-load}. Pruning vocabulary can greatly reduce the model size for multilingual PLMs.

In this paper, we present TextPruner, a model pruning toolkit for PLMs. It combines both transformer pruning and vocabulary pruning. The purpose of TextPruner is to offer a universal, fast, and easy-to-use tool for model compression. We expect it can be accessible to users with little model training experience. Therefore, we implement the structured optimization-free pruning methods for its convenient use and fast computation. Pruning a base-sized model only requires several minutes with TextPruner. TextPruner can also be a useful analysis tool for inspecting the importance of the neurons in the model.

TextPruner has the following highlights:
\begin{itemize}[noitemsep,topsep=0pt]
  \item TextPruner is designed to be easy to use. It provides both Python API and Command Line Interface (CLI). Working with either of them requires only a couple of lines of simple code. Besides, TextPruner is non-intrusive and compatible with Transformers \cite{wolf-etal-2020-transformers}, which means users do not have to change their models that are built on the Transformers library. 
  \item TextPruner works with different models and tasks. It has been tested on tasks like text classification, machine reading comprehension (MRC), named entity recognition (NER). TextPruner is also designed to be extensible for other models.
  \item TextPruner is flexible. Users can control the pruning process and explore pruning strategies via tuning the configurations to find the optimal configurations for the specific tasks.
\end{itemize}

\section{Pruning Methodology}
We briefly recall the multi-head attention (MHA) and the feed-forward network (FFN) in the transformers \cite{DBLP:conf/nips/VaswaniSPUJGKP17}. Then we describe how we prune the attention heads and the FFN based on the importance scores.

\subsection{MHA and FFN}
Suppose the input to a transformer is $\bm{X}\in \mathbb{R}^{n\times d}$ where $n$ is the sequence length and $d$ is the hidden size. the MHA layer with $N_h$ heads is parameterized by $\bm{W}^Q_i,\bm{W}^K_i,\bm{W}^V_i, \bm{W}^O_i \in \mathbb{R}^{d_h\times d}$
\begin{equation}
\mathrm{MHA}(\bm{X}) = \sum_i^{N_h} \mathrm{Att}_{\bm{W}^Q_i,\bm{W}^K_i,\bm{W}^V_i, \bm{W}^O_i}(\bm{X})
\end{equation}
where $d_h= d / N_h$ is the hidden size of each head. $\mathrm{Att}_{\bm{W}^Q_i,\bm{W}^K_i,\bm{W}^V_i, \bm{W}^O_i}(\bm{X})$ is the bilinear self-attention
\begin{flalign}
  & \mathrm{Att}_{\bm{W}^Q_i,\bm{W}^K_i,\bm{W}^V_i, \bm{W}^O_i}(\bm{X}) = \nonumber \\
  & \mathrm{softmax}(\frac{\bm{X}(\bm{W}^Q_i)^\top\bm{W}^K_i\bm{X}^\top}{\sqrt{d}})\bm{X}(\bm{W}^V_i)^\top \bm{W}^O_i
\end{flalign}

Each transformer contains a fully connected feed-forward network (FFN) following MHA. It consists of two linear transformations with a GeLU activation in between
\begin{flalign}
 \textrm{FFN}&_{\bm{W}_1,\bm{b}_1,\bm{W}_2,\bm{b}_2}(\bm{X}) = \nonumber \\
& \textrm{GeLU}(\bm{X} \bm{W}_1 + \bm{b}_1)\bm{W_2} + \bm{b}_2
\end{flalign}
where $\bm{W}_1 \in \mathbb{R}^{d\times d_{ff}}$, $\bm{W}_2 \in \mathbb{R}^{d_{ff}\times d}$, $\bm{b}_1 \in \mathbb{R}^{d_{ff}}$, $\bm{b}_2 \in \mathbb{R}^{d}$. $d_{ff}$ is the FFN hidden size. The adding operations are broadcasted along the sequence length dimension $n$. 

\subsection{Pruning with Importance Scores}
With the hidden size fixed, The size of a transformer can be reduced by removing the attention heads or removing the intermediate neurons in the FFN layer (decreasing $d_{ff}$, which is mathematically equal to removing columns from $\bm{W}_1$ and rows from $\bm{W}_2$). Following \citet{DBLP:conf/nips/MichelLN19}, we sort all the attention heads and FFN neurons according to their proxy importance scores and then remove them iteratively.

A commonly used importance score is the sensitivity of the loss with respect to the values of the neurons. We denote a set of neurons or their outputs as $\bm\Theta$. Its importance score is computed by
\begin{equation}\label{ISeq}
  \mathrm{IS}(\bm\Theta) = \mathbb{E}_{x\sim X}\left\lvert \frac{\partial \mathcal{L}(x)}{\partial\bm\Theta}\bm\Theta \right\rvert
\end{equation}
The expression in the absolute sign is the first-order Taylor approximation of the loss $\mathcal{L}$ around $\Theta=0$. Taking $\bm\Theta$ to be the output of an attention head $h_i$, $\mathrm{IS}(\bm\Theta)$ gives the importance score of the head $i$; Taking $\bm\Theta$ to be the set of the $i$-th column of $\bm{W}_1$, $i$-the row of $\bm{W}_2$ and the $i$-th element of $\bm{b}_1$,  $\mathrm{IS}(\bm\Theta)$ gives the importance score of the $i$-th intermeidate neuron in the FFN layer.

A lower importance score means the loss is less sensitive to the neurons. Therefore, the neurons are pruned in the order of increasing scores. In practice, we use the development set or a subset of the training set to compute the importance score.

\subsection{Self-Supervied Pruning}
In equation \eqref{ISeq}, the loss $\mathcal{L}$ usually is the training loss. However, there can be other choices of $\mathcal{L}$. We propose to use the Kullback–Leibler divergence to measure the varitaion of the model outputs:
\begin{equation}\label{ISkl}
  \mathcal{L}_{\mathrm{KL}}(x)=\mathrm{KL}(\mathrm{stopgrad}(q(x))||p(x))
\end{equation}
where $q(x)$ is the original model prediction distribution and $p(x)$ is the to-be-pruned model prediction distribution. The \texttt{stopgrad} operation is used to stop back-propagating gradients.
An increase in $\mathcal{L}_{\mathrm{KL}}$ indicates an increase in the diviation of $p(x)$ from the original prediction $q(x)$. 
Thus the gradient of $\mathcal{L}_{\mathrm{KL}}$ reflects the sensitivity of the model to the value of the neurons.
Evaluation of $\mathcal{L}_{\mathrm{KL}}$ does not require label information. Therefore the pruning process can be performed in a self-supervised way where the unpruned model provides the soft-labels $q(x)$. We call the method \textit{self-supervised pruning}. TextPruner supports both supervised pruning (where $\mathcal{L}$ is the training loss) and self-supervised pruning. We will compare them in the experiments.

\section{Overview of TextPruner}

\begin{figure}[!tbp]
  \centering
   \includegraphics[width=\linewidth]{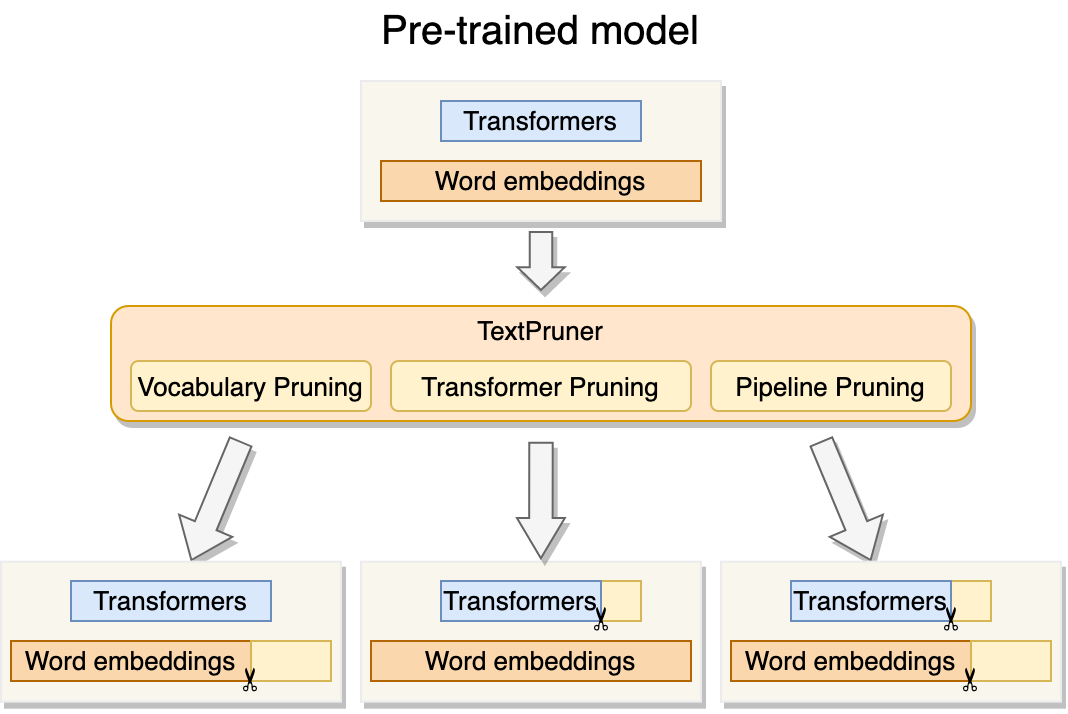} 
  \caption{\label{PruningModes} Three pruning modes in TextPruner.}
 \end{figure}

\subsection{Pruning Mode}

As illustrated in Figure \ref{PruningModes}, there are three pruning modes In TextPruner.

\paragraph{Vocabulary Pruning} 
The pre-trained models have a large vocabulary, but some tokens in the vocabulary rarely appear in the downstream tasks.
These tokens can be removed to reduce the model size and accelerate the training speed of the tasks that require predicting probabilities over the whole vocabulary.
In this mode, TextPruner reads and tokenizes an input corpus. TextPruner goes through the vocabulary and checks if the token in the vocabulary has appeared in the text file. If not, the token will be removed from both the model's embedding matrix and the tokenizer's vocabulary.

\paragraph{Transformer Pruning}
Previous studies \cite{DBLP:conf/nips/MichelLN19,voita-etal-2019-analyzing} have shown that not all attention heads are equally important in the transformers, and some of the attention heads can be pruned without performance loss \citep{cui-etal-2022}.
Thus, Identifying and removing the least important attention heads can reduce the model size and have a small impact on performance.

In this mode, TextPruner reads the examples and computes the importance scores of attention heads and the feed-forward networks' neurons. The heads and the neurons with the lowest scores are removed first. This process is repeated until the model has been reduced to the target size. TextPruner also supports custom pruning from user-provided masks without computing the importance scores.
\paragraph{Pipeline Pruning}
In this mode, TextPruner performs transformer pruning and vocabulary pruning automatically to fully reduce the model size.

\subsection{Pruners}

\begin{figure*}[!tbp]
  \centering
   \includegraphics[width=0.9\linewidth]{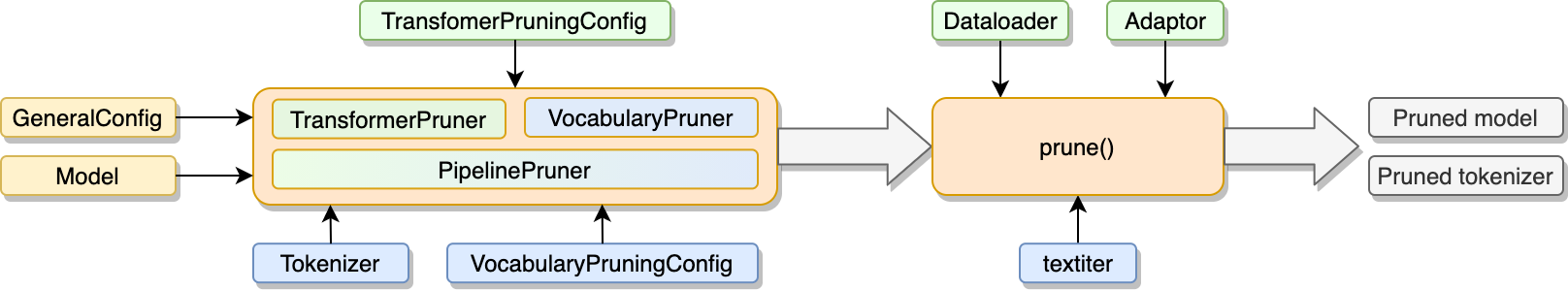} 
  \caption{\label{workflow} The workflow of TextPruner. The yellow blocks are the general arguments for any pruners. The green blocks should be provided for the TransformerPruner and PipelinePruner. The blue blocks should be provided for the VocabularyPruner and PipelinePruner.}
 \end{figure*}

The pruners are the cores of TextPruner, and they perform the actual pruning process.
There are three pruner classes, corresponding to the three aforementioned pruning modes: \textbf{VocabularyPruner}, \textbf{TransformerPruner}  and \textbf{PipelinePruner}. 
 Once the pruner is intialized, call the \texttt{pruner.prune($\ldots$)} to start pruning.

\subsection{Configurations}
The following configuration objects set the pruning strategies and the experiment settings.

\paragraph{GeneralConfig} It sets the device to use (CPU or CUDA) and the output directory for model saving.

\paragraph{VocabularyPruningConfig} It sets the token pruning threshold \texttt{min\_count} and whether pruning the LM head \texttt{prune\_lm\_head}. The token is to be removed from the vocabulary if it appears less than \texttt{min\_count} times in the corpus; if \texttt{prune\_lm\_head} is true, TextPruner prunes the linear transformation in the LM head too.

\paragraph{TransformerPruningConfig} The transformer pruning parameters include but not are limited to:
\begin{itemize}[noitemsep,topsep=0pt]
  \item \texttt{pruning\_method} can be \textit{mask} or \textit{iterative}. If it is \textit{iterative}, the pruner prunes the model based on the importance scores; if it is \textit{mask}, the pruner prunes the model with the masks given by the users.
  \item \texttt{target\_ffn\_size} denotes the average FFN hidden size $d_{ff}$ per layer.
  \item \texttt{target\_num\_of\_heads} denotes the average number of attention heads per layer.
  \item \texttt{n\_iters} is number of pruning iterations. For example, if the original model has $N_h$ heads per layer, the target model has $N'_h$ heads per layer, the pruner will prune $(N_h-N'_h)/\mathit{n\_iters}$ heads on average per layer per iteration. It also applies to the FFN neurons.
  \item If \texttt{ffn\_even\_masking} is true, all the FFN layers are pruned to the same size $d_{ff}$; otherwise, the FFN sizes vary from layer to layer and their average size is $d_{ff}$.
  \item If \texttt{head\_even\_masking} is true, all the MHAs are pruned to the same number of heads; otherwise, the number of attention heads varies from layer to layer.
  \item If \texttt{ffn\_even\_masking} is false, the FFN hidden size of each layer is restricted to be a multiple of \texttt{multiple\_of}. It make the model structure friendly to the device that works most efficiently when the matrix shapes are multiple of a specific size.
  \item If \texttt{use\_logits} is true, self-supervised pruning is enabled.
\end{itemize}

\begin{figure}[!t]
  \centering
   \includegraphics[width=\linewidth]{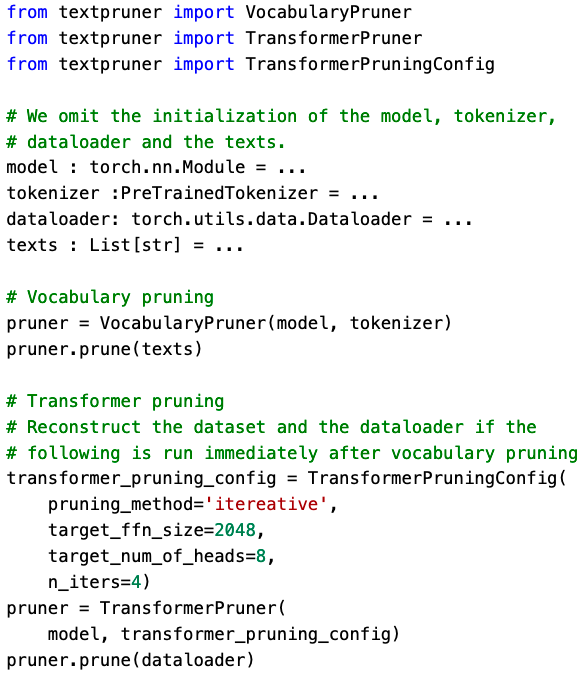} 
  \caption{\label{workflowExample} A typical TextPruner workflow for transformer pruning and vocabulary pruning.}
 \end{figure}

All the configurations can be initialized manually in python scripts or from JSON files (for the CLI, the configurations can only be initialized from the JSON files).
An example of the configuration in a Python script is shown in Figure \ref{workflowExample}.

 \begin{table*}[]
  \small
  \centering
  \begin{tabular}{@{}lcccccc@{}}
    \toprule
    Model                             & Vocabulary size & Model size & Dev (en) & Dev (zh) & Test (en) & Test (zh) \\ \midrule
    XLM-R                              & 250002         & 1060 MB ($100\%$)    & 84.8     & 75.1     & 85.7      & 75.0      \\ \midrule
    \ + Vocabulary Pruning on en        & 26653           & 406 MB ($38.3\%$)    & 84.6     & -        & 85.9      & -         \\
    \ + Vocabulary Pruning on zh        & 23553           & 397 MB ($37.5\%$)    & -        & 74.7     & -         & 74.5      \\
    \ + Vocabulary Pruning on en and zh & 37503           & 438 MB ($41.3\%$)    & 84.8     & 74.3     & 85.8      & 74.5      \\ \bottomrule
    \end{tabular}
    \caption{\label{XNLIVocab} The accuracy scores ($\times 100\%$) of models with the pruned vocabulary on XNLI dev set and test set.}
  \end{table*}

\subsection{Other utilities}
TextPruner contains diagnostic tools such as \textbf{summary} which inspects and counts the model parameters, and \textbf{inference\_time} which measures the model inference speed. Readers may refer to the examples in the repository to see their usages.

\subsection{Usage and Workflow}\label{usageandworkflow}
TextPruner provides both Python API and CLI.
The typical workflow is shown in Figure \ref{workflow}. 
Before calling or Initializing TextPruner, users should prepare:
\begin{enumerate}[noitemsep,topsep=0pt]
  \item A trained a model that needs to be pruned. 
  \item For vocabulary pruning, a text file that defines the new vocabulary.
  \item For transformer pruning, a python script file that defines a dataloader and an adaptor. 
  \item For pipeline pruning, both the text file and the python script file.
\end{enumerate}

\paragraph{Adaptor} It is a user-defined function that takes the model outputs as the argument and returns the loss or logits. It is responsible for interpreting the model outputs for the pruner. If the adaptor is \texttt{None}, the pruner will try to infer the loss from the model outputs.

\paragraph{Pruning with Python API}
First, initialize the configurations and the pruner, then call \texttt{pruner.prune}  with the required arguments, as shown in Figure \ref{workflow}. Figure \ref{workflowExample} shows an example. Note that we have not constructed the GeneralConfig and VocabularyPruningConfig. The pruners will use the default configurations if they are not specified, which simplifies the coding.

\paragraph{Pruning with CLI}
First create the configuration JSON files, then run the \texttt{textpruner-cli}. Pipeline pruning example:
\begin{lstlisting}[language=bash]
textpruner-cli  \
  --pruning_mode pipeline \
  --configurations vocab.json trm.json \
  --model_class BertForClassification \
  --tokenizer_class BertTokenizer \
  --model_path models/ \
  --vocabulary texts.txt \
  --dataloader_and_adaptor dataloader.py
  \end{lstlisting}

\subsection{Computational Cost}
\paragraph{Vocabulary Pruning} The main computational cost in vocabulary pruning is tokenization. This process will take from a few minutes to tens of minutes, depending on the corpus size. However, the computational cost is negligible if the pre-tokenized text is provided.

\paragraph{Transformer Pruning} The main computational cost in transformer pruning is the calculation of importance scores. It involves forward and backward propagation of the dataset. This cost is proportional to $\texttt{n\_iters}$ and dataset size. As will be shown in Section \ref{ex_results}, in a typical classification task, a dataset with a few thousand examples and setting $\texttt{n\_iters}$ around $10$ can lead to a decent performance. This process usually takes several minutes on a modern GPU (e.g., Nvidia V100).

\subsection{Extensibility}
TextPruner supports different pre-trained models and the tokenizers via the model structure definitions and the tokenizer helper functions registered in the \texttt{MODEL\_MAP} dictionary.
Updating TextPruner for supporting more pre-trained models is easy. Users need to write a model structure definition and register it to the \texttt{MODEL\_MAP}, so that the pruners can recognize the new model.

\begin{table}[t!]
  \small
  \centering
  \begin{tabular}{@{}c|cccc@{}}
  \toprule
  Structure & 12    & 10   & 8    & 6    \\ \midrule
  3072       & \makecell[c]{100\% \\ (1.00x)} & \makecell[c]{89\% \\ (1.08x) }& \makecell[c]{78\% \\ (1.19x) }& \makecell[c]{67\% \\ (1.30x) }\\
  2560       & \makecell[c]{94\% \\ (1.08x) } & \makecell[c]{83\% \\ (1.18x) }& \makecell[c]{72\% \\ (1.29x) }& \makecell[c]{61\% \\ (1.44x) }\\
  2048       & \makecell[c]{89\% \\ (1.17x) } & \makecell[c]{78\% \\ (1.28x) }& \makecell[c]{67\% \\ (1.43x) }& \makecell[c]{56\% \\ (1.63x) }\\
  1536       & \makecell[c]{83\% \\ (1.29x) } & \makecell[c]{72\% \\ (1.42x) }& \makecell[c]{61\% \\ (1.63x) }& \makecell[c]{50\% \\ (1.90x) }\\ \bottomrule
  \end{tabular}
  \caption{\label{size_and_speed}Transformer sizes (listed as percentages) and speedups (listed in the parentheses) of different structures relative to the base model $(12, 3072)$.}
  \end{table}

\section{Experiments}

In this section, we conduct several experiments to show TextPruner's ability to prune different pre-trained models on different NLP tasks.
We mainly focus on the text classification task. We list the results on the MRC task and NER task with different pre-trained models in the Appendix.

\subsection{Dataset and Model}
We use the Cross-lingual Natural Language Inference (XNLI) corpus \cite{conneau2018xnli} as the text classification dataset and build the classification model based on XLM-RoBERTa \cite{conneau-etal-2020-unsupervised}.
The model is \textit{base}-sized with 12 transformer layers with  FFN size 3072, hidden size 768, and 12 attention heads per layer. Since XNLI is a multilingual dataset, we fine-tune the XLM-R model on the English training set and test it on the English and Chinese test sets to evaluate both the in-language and zero-shot performance. 

\subsection{Results on Text Classification}\label{ex_results}

\paragraph{Effects of Vocabulary Pruning}
As XLM-R is a multilingual model, We conduct vocabulary pruning on XLM-R with different languages, as shown in Table \ref{XNLIVocab}. We prune XLM-R on the training set of each language, i.e., we only keep the tokens that appear in the training set. 

When pruning on the English and Chinese training sets separately,
the performance drops slightly . After pruning on both training sets, the model size still can be greatly reduced by about $60\%$ while keeping a decent performance.

Vocabulary pruning is an effective method for reducing multilingual pre-trained model size, and it is especially suitable for tailoring the multilingual model for specific languages.

\begin{figure}[t!]
  \centering
   \includegraphics[width=\linewidth]{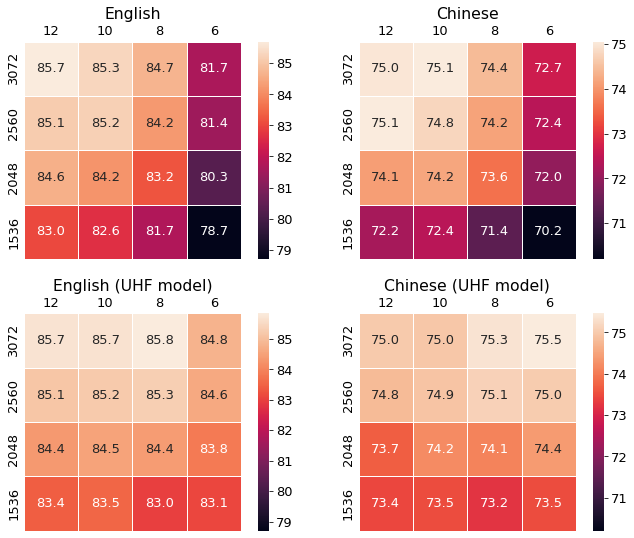}
  \caption{\label{heatmap} The Performance of the pruned models with different structures on the test sets. The x-axis represents different average numbers of attention heads; the y-axis represents different average FFN sizes. Left column: the accuracy scores on the English test set; Right column: the accuracy scores on the Chinese test set. Models in the first row have homogenous structures, while models in the second row do not. {UHF} stands for uneven heads and FFN neurons.}
 \end{figure}

\paragraph{Effects of Transformer Pruning}
For simplicity, we use the notation $(H, F)$ to denote the model structure, where $H$ is the average number of attention heads per layer, $F$ is the average FFN hidden size per layer. With this notation, the original (unpruned) model is $(12, 3072)$. Before we show the results on the specific task, we list the transformer sizes and their speedups of different target structures relative to the unpruned model $(12, 3072)$ in the Table \ref{size_and_speed}.

We compute the importance scores on the English development set. The number of iterations $n_{iters}$ is set to 16. We report the mean accuracy of five runs. The performance on English and Chinese test sets are shown in Figure \ref{heatmap}. 
The top-left corner of each heatmap represents the performance of the original model. The bottom right corner represents the model $(6,1536)$, which contains half attention heads and half FFN neurons.

The models in heatmaps from the first row have homogenous structures: each transformer in the model has the same number of attention heads and same FFN size, while the models in the bottom heatmaps have uneven numbers of attention heads and FFN sizes in transformers. We use the abbreviation \textit{UHF} (\textbf{U}neven \textbf{H}eads and \textbf{F}FN neurons) to distinguish them from homogenous structures. 
We see that by allowing each transformer to have different sizes, the pruner has more freedom to choose the neurons to prune, thus the UHF models perform better than the homogenous ones.

Note that the model is fine-tuned on the English dataset. The performance on Chinese is zero-shot. After pruning on the English development set, the drops in the performance on Chinese are not larger than the drops in the performance on English. It means the important neurons for the Chinese task remain in the pruned model. In the multilingual model, the neurons that deal with semantic understanding do not specialize in specific languages but provide cross-lingual understanding abilities. 

\begin{figure}[!tbp]
  \centering
   \includegraphics[width=\linewidth]{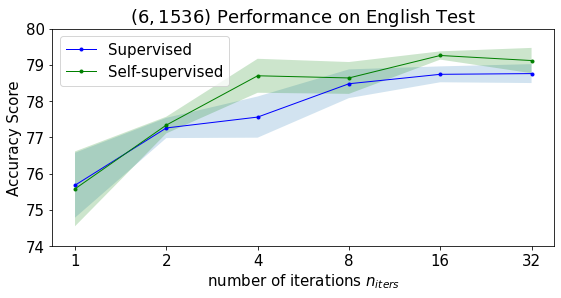}
   \includegraphics[width=\linewidth]{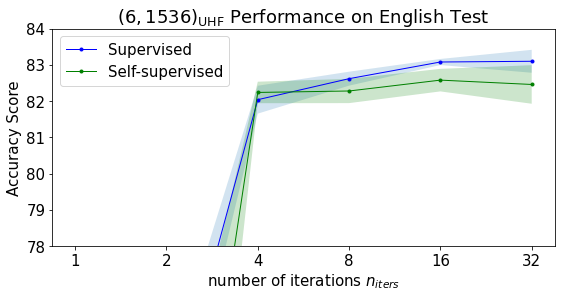} 
  \caption{\label{n_iters} Model Performance on the English test set with different number of iterations.}
 \end{figure}

Figure \ref{n_iters} shows how $n_{iters}$ affects the performance.
We inspect both the non-UHF model $(6, 1536)$ and the UHF model $(6, 1536)_{\mathrm{UHF}}$.
The solid lines denote the average performance over the five runs.
The shadowed area denotes the standard deviation.
In all cases, the performance grows with the $n_{iters}$. 
Pruning with only one iteration is a bad choice and leads to very low scores. 
We suggest setting $n_{iters}$ to at least 8 for good enough performance.

In Figure \ref{n_iters} we also compare the supervised pruning (with $\mathcal{L}$ being the cross-entropy loss with the ground-truth labels) and the proposed self-supervised pruning (with $\mathcal{L}$ being the KL-divergence Eq \eqref{ISkl}) . Although no label information is available, the self-supervised method achieves comparable and sometimes even higher results.

\begin{figure}[!tbp]
  \centering
   \includegraphics[width=\linewidth]{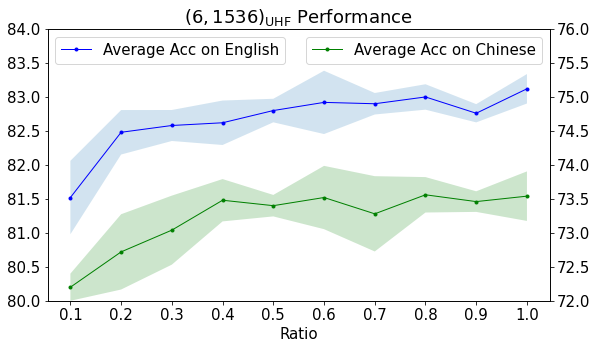}
  \caption{\label{ratio_pic} Model Performance on the test set with different number of examples for computing importance scores. Left y-axis: accuracy on Enligh. Right y-axis: accuracy on Chinese.}
 \end{figure}

How much data are needed for model pruning? To answer this question, we randomly sample $10\%$, $20\%$, $\ldots$, $90\%$, $100\%$ examples from the English development set for computing importance scores. We inspect the $(6,1536)_{\mathrm{UHF}}$ model. Each experiment has been run five times.
The results are shown in Figure \ref{ratio_pic}. With about $70\%$ examples (about 1.7K examples) from the development set, the pruned model achieves a performance that is nearly comparable with the model pruned with the full development set (2490 examples).

\section{Conclusion and Future Work}
This paper presents TextPruner, a model pruning toolkit for pre-trained models. It leverages optimization-free pruning methods, including vocabulary pruning and transformer pruning to reduce the model size. It provides rich configuration options for users to explore and experiment with. TextPruner is suitable for users who want to prune their model quickly and easily, and it can also be used for analyzing pre-trained models by pruning, as we did in the experiments.

For future work, we will update TextPruner to support more pre-trained models, such as the generation model T5 \cite{2020t5}. We also plan to combine TextPruner with our previously released knowledge distillation toolkit TextBrewer \cite{textbrewer-acl2020-demo} into a single framework to provide more effective model compression methods and a uniform interface for knowledge distillation and model pruning.

\section*{Acknowledgements}
This work is supported by the National Key Research and Development Program of China via grant No. 2018YFB1005100.

\bibliography{custom}
\bibliographystyle{acl_natbib}

\appendix

\section{Datasets and Models}
We experiment with different pre-trained models to test TextPruner's ability to prune different models.
For the MRC task, we use SQuAD \cite{rajpurkar-etal-2016-squad} dataset and RoBERTa \cite{liu2019roberta} model;
For the NER task, we use CoNLL 2003 \cite{tjong-kim-sang-de-meulder-2003-introduction} and BERT \cite{devlin-etal-2019-bert} model.
All the models are \textit{base}-sized, i.e., 12 transformer layers with a hidden size of 768, an FFN size of 3072, and 12 attention heads per layer.

\section{Transformer Pruning on MRC}
We compute the importance scores on a subset of the training set (5120 examples).
The F1 score on the SQuAD development set is listed in Table \ref{resultssquad}. $(12,3072)$ is the unpruned model. The performance grows with the $n_{iters}$. The number of iterations also plays an important role on model performance in the SQuAD task.
We also see that pruning with only one iteration is a bad choice and leads to low scores.  Setting $n_{iters}$ to at least 8 achieves good enough performance.

\section{Transformer Pruning on NER}
We compute the importance scores on the CoNLL 2003 development set.
The F1 score on the test is listed in Table \ref{resultsner}. We also see large gaps in performance between $n_{iters}=4$ and $n_{iters}=8$.

The performance of the pruned models with different structures is shown in Figure \ref{heatmap}. We only consider the UHF case for it can achieve the best overall performance. The number of iterations $n_{iters}$ is set to 16.

\begin{table}[t!]
    \centering
    \small
    \begin{tabular}{@{}lccccc@{}}
      \toprule
      Model           & 1     & 2     & 4     & 8     & 16    \\ \midrule
      $(12, 3072)$    & \multicolumn{5}{l}{91.4}             \\ \midrule
      $(8 , 2048)$    & 76.4 & 80.3 & 81.9 & \textbf{82.9} & 82.5 \\
      $(8 , 2048)_{\mathrm{UHF}}$ & 87.5  &  86.4    & 87.6   & 88.3      &  \textbf{88.4}     \\
      
      $(6, 1536)$     & 12.8      &  42.6   &  49.5     &  51.5     &   \textbf{56.5}    \\
      $(6 , 1536)_{\mathrm{UHF}}$ & 47.2      & 55.6  & 66.1  &  74.1 & \textbf{75.2}  \\ \bottomrule
      \end{tabular}
      \caption{\label{resultssquad} The F1 score on SQuAD. Each score is averaged over five runs. Different columns represent results under different number of iterations. We \textbf{bold} the best F1 in each row.}
      \end{table}
  
      \begin{table}[t!]
        \centering
        \small
        \begin{tabular}{@{}lcccccc@{}}
          \toprule
          Model           & 1     & 2     & 4     & 8     & 16 & 32   \\ \midrule
          $(12, 3072)$    & \multicolumn{6}{l}{91.3}             \\ \midrule
          $(8 , 2048)$    & 88.5 & 88.4 & 88.7 & 89.2 & 89.2 & 89.4 \\
          $(8 , 2048)_{\mathrm{UHF}}$ & 81.8  &  90.0   & 90.6  & 90.7    & 90.8 & 90.8   \\
          
          $(6, 1536)$     & 33.6   &  56.2  & 62.4   & 80.5   & 83.4  & 84.1 \\
          $(6 , 1536)_{\mathrm{UHF}}$ & 9.8  & 67.6  & 80.2  & 86.2 & 87.0 & 87.3 \\ \bottomrule
          \end{tabular}
          \caption{\label{resultsner} The F1 score on CoNLL 2003. Each score is averaged over five runs.}
          \end{table}

\begin{figure}[t!]
\centering
    \includegraphics[width=0.8\linewidth]{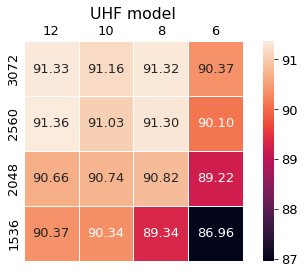}
\caption{\label{heatmap} The Performance of the pruned models with different structures on the CoNLL 2003 test set. Each score is averaged over five runs.}
\end{figure}

\end{document}